\documentclass[graybox,envcountchap]{SNmult}

\usepackage{type1cm}
\usepackage{makeidx}
\usepackage{graphicx}
\usepackage{multicol}
\usepackage[bottom]{footmisc}
\usepackage{newtxtext}
\usepackage[varvw]{newtxmath}

\usepackage[utf8]{inputenc}
\usepackage{booktabs}
\usepackage[numbers,sort&compress,comma,square,sectionbib]{natbib}
\usepackage{chapterbib}
\usepackage{doi}
\usepackage[dvipsnames]{xcolor}
\usepackage{subcaption}
\usepackage[thinc]{esdiff}
\usepackage{float}

\newcommand{\secref}[1]{Section~\ref{#1}}

\newcommand{\figref}[1]{Figure~\ref{#1}}
\newcommand{\tabref}[1]{Table~\ref{#1}}

\definecolor{recommendation}{gray}{0.8}

\definecolor{exploration}{gray}{0.8}
\newenvironment{exploration}[1]{\ignorespaces\def\stmtopen##1{##1}%
\formtmp{exploration}{\ \circledmark{\textcolor{black}{$~\blacktriangleright$}}{\kern6pt}#1}}{\par\noindent\textcolor{exploration}{\rule{\columnwidth}{1pt}}\vskip2pt\par\addvspace{\baselineskip}}%

\makeindex

\begin{document}

\title*{Developing AI Agents with Simulated Data}
\subtitle{Why, what, and how?}

\author{Xiaoran Liu and Istvan David}

\institute{Xiaoran Liu \at McMaster University, Canada\\
\email{liu2706@mcmaster.ca}
\and
Istvan David \at McSCert, McMaster University, Canada\\ \email{istvan.david@mcmaster.ca}}

\maketitle

\abstract{As insufficient data volume and quality remain the key impediments to the adoption of modern subsymbolic AI, techniques of synthetic data generation are in high demand. Simulation offers an apt, systematic approach to generating diverse synthetic data. This chapter introduces the reader to the key concepts, benefits, and challenges of simulation-based synthetic data generation for AI training purposes, and to a reference framework to describe, design, and analyze digital twin-based AI simulation solutions.}

\section{Introduction}

Modern artificial intelligence (AI) is enabled by large volumes of high-quality data, and the capacity to process it using powerful computational techniques~\cite{zhou2017machine}. In modern AI, the performance and reliability of AI models are closely linked to the characteristics of their training data~\cite{rujas2025synthetic}. This is a stark contrast with classical symbolic AI, supported by formal methods and logic~\cite{smolensky1987connectionist}.
However, acquiring data that is reliably of high volume and capacity, is often a challenge. Barriers, such as acquisition costs, privacy constraints, and safety concerns often limit access to real-world data~\cite{zhou2017machine}. In many engineering domains, proprietary data, data silos, and sensitive operational procedures complicate the acquisition of data~\cite{farahani2023smart}. The acute scarcity of labeled data, a precondition for many supervised learning tasks, is also well-documented~\cite{hagendorff202015}. And even when data is available, issues such as incompleteness, inconsistencies, duplication, and noise can compromise its usability.

To address these limitations associated with real-world data, various approaches for generating synthetic data have been proposed~\cite{figueira2022survey}.
Synthetic data is generated from a sufficiently detailed and realistic model. Synthetic data can be used to enrich real data or to completely replace it~\cite{jordon2022synthetic}. The appeal of such techniques is obvious as generating synthetic data has much lower costs than obtaining real data, e.g., through measurements and observations that may be costly, time-consuming, and hazardous.

A particular way of generating synthetic data is simulation. Simulators are programs that encode a system's probabilistic behavior and enact it to computationally exhibit various states of the system~\cite{ross2022simulation}. This enactment is called the simulation and its output, the simulation trace, reflects system's behavior and constitutes the data that can be used for AI training.
By generating valuable data in a virtual environment, simulation offers a scalable and cost-effective alternative to real-world data collection. \index{AI simulation|(}Recently, Garner, one of the leading consulting firms identified ``AI simulation'' as a fast-emerging technological trend. In their definition, AI simulation is ``\textit{the combined application of AI and simulation technologies to jointly develop AI agents and the simulated environments in which they can be trained, tested and sometimes deployed. It includes both the use of AI to make simulations more efficient and useful, and the use of a wide range of simulation models to develop more versatile and adaptive AI systems}''~\cite{aisim-gartner}. Clearly, simulation-based methods for producing AI training data are drawing significant attention from researchers, practitioners, and companies alike.\index{AI simulation|)}
For example, \citet{bu2021carla} use the CARLA~\cite{dosovitskiy2017carla} simulator to generate diverse synthetic images of rare street objects (e.g., fire hydrants and crosswalks) with automatic annotations, addressing the data scarcity challenge in object detection.
\citet{marcinandrychowicz2020learning} use simulated environments based on the MuJoCo~\cite{todorov2012mujoco} physics engine to train reinforcement learning (RL) policies for dexterous robotic hand manipulation, successfully transferring emergent human-like manipulating behaviors to the physical robot.

This chapter introduces the reader to the key concepts, benefits, and challenges of simulation-based AI training data generation, and to a reference framework to describe, design, and analyze digital twin-based AI simulation solutions.

\section{Simulating Data for Training AI}\label{sec:sim4ai}

We now delve into the foundations of simulating data for AI training purposes, the typical techniques to do so, and the respective challenges.

\subsection{Simulation and data}

A simulation is an imitative representation of a system that could exist in the real world~\cite {ross2022simulation}. A simulator (cf. simulation) is a program that implements the probabilistic mechanisms that are considered the simulation of the real-world phenomenon~\cite{zeigler2018theory}.

\begin{figure}
    \centering
    \includegraphics[width=0.75\linewidth]{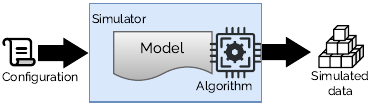}
    \caption{Schematic overview of AI training data generation by simulation}
    \label{fig:schema}
\end{figure}

After modeling the phenomenon or system, a simulation of the model computes the dynamic behavior that represent the system~\cite{vangheluwe2002introduction}. The execution of a simulation produces a \textit{simulation trace}, an output that represents how the simulated system behaves over time~\cite{ross2022simulation}. As shown in \figref{fig:schema}, these simulation traces constitute the \textit{data} that can be used to train or tune AI agents. Simulated data is cheaper and faster to generate~\cite{muratore2022robot}, and provides flexible replication of complex real-world scenarios, enabling controlled experimentation and reliable algorithm testing~\cite{choi2021use}.

\begin{figure}
    \centering
    \includegraphics[width=0.5\linewidth]{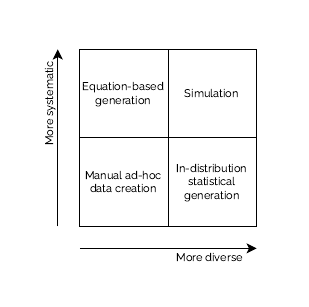}
    \caption{Typical data generation techniques for AI training}
    \label{fig:methods}
\end{figure}

There are various ways of producing synthetic data for AI training, but simulation uniquely combines data diversity with a systematic approach, as shown in \figref{fig:methods}. Manual ad-hoc data generation, e.g, by filling in missing data points by an expert is the simplest way of generating synthetic data. Apart from not scaling well to the data volume needs of AI, such methods are also informal, not repeatable or systematic, and the generated data typically lacks diversity. Equation-based methods can generate data in a systematic, repeatable fashion, but lack diversity. For example, generating positional data from elementary equations of kinematics, e.g., $a(t)=\diff[2]{s}{t}$. The result of numerically solving this equation will always be the same---however, training AI requires some diversity.
To address this need, statistics-based methods can be used to generate data points that still adhere to a distribution and preserve statistical properties, e.g., of a pre-existing dataset.
Simulation improves over the previous techniques by combining the power of systematic techniques with diversity in the generated data.

\subsection{Simulation methods for data generation}

A variety of simulation methods are available to generate AI training data. In the following, we review some them. This is, of course, far from being an exhaustive list, but should inform the reader and give some ideas. 

\subsubsection{Discrete simulation}\index{simulation!discrete simulation|(}
Discrete simulation models have variables whose values vary only at discrete points in time~\cite{ozgun2009discrete}, making them particularly suitable for logistics~\cite{agalianos2020discrete}, healthcare~\cite{brailsford2001comparison}, and network systems~\cite{antoine2008discrete}. Discrete simulations demonstrate technical and practical potentialities that make them well-suited for detailed analysis of specific and well-defined linear systems~\cite{sweetser1999comparison}.

\index{Discrete Event Simulation (DES)|(}
\textit{Discrete-event simulation (DES)} represents a system's behavior as a sequence of events occurring at specific points in time, where each event triggers a change in the system's state~\cite{varga2001discrete}. For example, \citet{chan2022generation} use DES to generate production data in various manufacturing scenarios, providing labeled synthetic data for training machine learning models.\index{Discrete Event Simulation (DES)|)}

\index{Agent-based model (ABM)|(}
\textit{Agent-based simulation (ABS)}, also known as agent-based modeling (ABM), is a computational approach used to simulate dynamic processes composed of independent, autonomous agents~\cite{macal2014introductory}. ABS is most frequently employed to simulate individual decision-making and patterns of behavior within social or organizational contexts~\cite{bonabeau2002agent}. For example, \citet{lombardo2022unsupervised} use synthetic data generated by agent-based simulation to train an unsupervised deep recurrent neural network, to distinguish anomaly trajectories depending on the users' roles within a building.\index{Agent-based model (ABM)|)}

Discrete simulation provides a compact model representation, regardless of complexity, and avoids the state explosion problems common in Markov models~\cite{siebert2012state}. However, discrete simulation cannot address system bounded input-bounded output (BIBO) stability, which is essential for strategic planning and aggregated decision-making~\cite{brito2011conceptual}. Also, when macroscopic or long-term strategic approaches are required, discrete simulation involves building overly complex models incompatible with global perspectives~\cite{lin1998generic}. Moreover, discrete models typically offer limited insights into the underlying dynamics between system variables, resulting in reduced application capacity in systems with elevated dynamic complexity~\cite{brito2011conceptual}.\index{simulation!discrete simulation|)}

\subsubsection{Continuous simulation}\index{simulation!continuous simulation|(}
Continuous simulation captures systems whose variables change continuously over time, typically governed by differential equations, as seen in chemical processes or thermal systems~\cite{cellier2013continuous}, design flood estimation~\cite{boughton2003continuous}, and electromechanical dynamics in power systems~\cite{parashar2004continuum}.

\index{System Dynamics (SD)|(}
\textit{System dynamics (SD) simulation} models complex system behavior over time using feedback loops, stock-and-flow structures, and time delays~\cite{mustafee2010profiling}. It excels in capturing the nonlinear dynamics of complex systems, where objects flow between stocks and accumulate over time~\cite{davahli2020system}. System dynamics effectively supports strategic decision-making through its integrative perspective at an elevated managerial level~\cite{helal2008hybrid}. The high degree of variable aggregation and explicit causal relationships focus attention on system dynamic behavior, making system dynamics models valuable tools for management strategy analysis in organizationally complex problems~\cite{brito2011conceptual}. This approach is particularly useful for policy strategy analysis by incorporating soft behavioral aspects that, while difficult to quantify, significantly affect system performance~\cite{sweetser1999comparison}.
For example, \citet{roozkhosh2023blockchain} generate synthetic data in 10 years by using system dynamics simulation, modeling how key factors like inflation, exchange rate, policy impact affect blockchain acceptance in supply chains. This data is used to train neural networks for long-term prediction. \index{System Dynamics (SD)|)}

\index{computational fluid dynamics (CFD)|(}
\textit{Computational fluid dynamics (CFD) simulation} is a numerical approach for simulating fluid flow based on fundamental physical laws: conservation of mass, conservation of energy, and Newton’s second law~\cite{bhatti2020recent}. These laws are expressed as partial differential equations that describe fluid behavior within a given domain. Computational fluid dynamics excels in detailed fluid behavior analysis and visualization~\cite{bhatti2020recent}.
For example, \citet{ashton2024windsorml} present the WindsorML dataset, which provides CFD-simulated aerodynamic data for Windsor body variants, enabling the training of machine learning surrogate models for automotive applications.
\index{computational fluid dynamics (CFD)|)}
\index{simulation!continuous simulation|)}

\subsubsection{Monte Carlo Simulation} \index{simulation!Monte Carlo|(}

Monte Carlo simulation is a method that uses random sampling and statistical modeling to estimate mathematical functions and mimic the operations of complex systems~\cite{harrison2010introduction}. It is particularly effective in domains where uncertainty and stochastic processes play a critical role~\cite{kumar2025understanding}.

In medical imaging, Monte Carlo methods are widely used to simulate realistic imaging data~\cite{ljungberg1989monte}. For example, \citet{leube2024deep} propose a deep learning–based correction method for medical imaging, using U-Net models trained on 10,000 synthetic scans generated through physics-based simulation. In a related study, they also trained U-Net models to recover missing image data from sparsely sampled scans, using another set of 10,000 simulated examples~\cite{leube2022analysis}.

In the supply chain domain, Monte Carlo simulation has been applied to generate training data under uncertainty~\cite{badakhshan2024application}. For example, \citet{vondra2019digestate} use Monte Carlo simulation to generate over 20,000 possible configurations to implement an evaporation system, and trained a neural network and a decision tree classifier to identify favorable decision pathways for energy consumption planning. \citet{sishi2021supply} use simulated data to train a regression model for energy optimization.

In dynamic pricing and revenue optimization, \citet{rana2014real} address the problem of real-time dynamic pricing under non-stationary demand using model-free reinforcement learning. They simulate customer interactions using Monte Carlo simulation to train and evaluate RL-based pricing policies. \index{simulation!Monte Carlo|)}

\subsubsection{Computer graphics-based simulation} \index{simulation!computer graphics|(}
Computer graphics-based simulation is a widely used approach for generating synthetic data, particularly for visual-based AI systems. It employs a computer graphics rendering pipeline to synthesize images based on manually created virtual assets, including 3D geometry, material properties, lighting, and motion parameters~\cite{xu2024systematic}. Advances in computer graphics tools, such as game engines like Unity~\footnote{https://unity.com/} and Unreal~\footnote{https://www.unrealengine.com/en-US}, have significantly improved the flexibility and realism of these pipelines. The increasing availability of high-quality 3D assets from public datasets and platforms (e.g., ShapeNet~\footnote{https://shapenet.org/}, Adobe Stock~\footnote{https://stock.adobe.com/ca/}, and Unity Asset Store~\footnote{https://assetstore.unity.com/}) has further made it easier to build such simulators.
In addition, visual and motion realism can be enhanced by increasing geometric complexity and incorporating advanced rendering algorithms or physics engines~\cite{pun2023neural, song2023synthetic}.
It has been successfully applied in tasks such as object detection~\cite{tremblay2018training}, pose estimation~\cite{ionescu2013human3}, and robot control~\cite{li2025evaluating}.

For example, CARLA is an open-source autonomous driving simulator built on Unreal Engine 4 (UE4), featuring custom-designed digital assets (e.g., vehicles, buildings, road layouts) that reflect real-world scales and properties~\cite{dosovitskiy2017carla}. It achieves high-quality synthetic data generation through state-of-the-art rendering and physics engines, including PhysXVehicles in UE4 for realistic vehicle dynamics. CARLA has been adopted for various AI training tasks. For perception, synthetic images rendered in CARLA have been used to augment low-shot street-view datasets, providing automatic annotations and improving object detection performance on real-world benchmarks~\cite{bu2021carla}. For control, CARLA serves as a safe testing environment for deep reinforcement learning (DRL) algorithms, allowing researchers to train vehicle control models that can navigate efficiently along predetermined routes~\cite{perez2022deep}.\index{simulation!computer graphics|)}

\section{Challenges in Developing AI Agents with Simulated Data}

Although simulators provide a controlled, safe, and cost-efficient way to train and test AI agents, there are some characteristic challenges that must be considered when developing simulation-based AI training. In this section, we review some of the associated challenges, with a special focus on the key challenge of the sim-to-real gap.

\subsection{The sim-to-real gap}\label{sec:sim2real}

\index{sim-to-real|(}\index{sim2real|(}When training AI models with simulators, a lot hinges on the quality, fidelity, and validity of the simulation models that are to capture the simulated phenomenon to sufficient detail.
Simulators are frequently built on idealized assumptions that may ignore seemingly minor but significant real-world factors such as object friction, air resistance, light changes, noise, and sensor delays, etc.~\cite{chebotar2019closing}.
When the predictions of the simulation model, i.e., its traces, do not align with real-world observations, the simulation model is said to exhibit a reality gap~\cite{kadian2020sim2real}, also called the sim-to-real gap~\cite{tobin2017domain}.
Training an AI with such a simulator allows the sim-to-real gap to propagate into the AI models, rendering them insufficient once deployed in a real system.

A major contributor to the sim-to-real gap is the simplification of complex phenomena during simulation. For example, lighting-aware digital twins often assume simplifications including approximate reconstruction, separate lighting prediction, and fixed base materials, which lead to imperfect scene reconstruction and degraded realism~\cite{pun2023neural}. 
Moreover, while simulation data is typically clean and well-structured, it often lacks the noise, variability, and rare events present in real-world data. Even when artificial noise is injected, it rarely captures the full complexity of real operational environments, raising concerns about generalization and long-term reliability~\cite{tamascelli2024artificial}.
The lack of sim-to-real considerations makes it challenging for AI agents to match their simulation-based performance in real-world settings.
Sim-to-real transfer is the procedure of transferring knowledge gained in simulated settings to real-world applications~\cite{hu2024how}. 
Effective sim-to-real transfer methods are essential for maximizing the benefits of simulation-based training, such as cost efficiency and safety, while ensuring that the policies perform well in real-world applications.\index{sim-to-real|)}\index{sim2real|)}

\subsubsection{Methods for sim-to-real mitigation}

In the following, we outline some methods to mitigate the sim-to-real gap.

\begin{description}
    \item[\textbf{Domain randomization}] \index{domain randomization|(}Domain randomization is a widely adopted technique to bridge the sim-to-real gap. Instead of relying on high-fidelity simulation,  it exposes AI models to varied simulation parameters during training, rather than training on a single simulated environment, to promote generalization~\cite{tobin2017domain}. 
    Thus, it enhances the model’s ability to operate reliably in real-world conditions~\cite{pitkevich2024survey}.

    \citet{zhao2020sim} categorize domain randomization techniques into two types based on which components of the simulator are randomized: \textit{visual} and \textit{dynamics} randomization. \textit{Visual domain randomization} randomizes elements of the visual scene such as lighting conditions, object textures, and backgrounds. For example, \citet{tremblay2018training} train object detectors using synthetic images with randomized lighting, textures, and poses, and the network yields better performance than using real data alone. \textit{Dynamics domain randomization} randomizes physical properties of the environment, such as object masses, joint frictions, and contact forces. For example, \citet{andrychowicz2020learning} train dexterous manipulation policies for a robotic hand by randomizing object sizes, masses, and friction coefficients in simulation, enabling successful transfer to the physical robot.\index{domain randomization|)}
    
    \item[\textbf{Domain adaptation}] \index{domain adaptation|(}Domain adaptation tackles the mismatch between two related domains: a \textit{source domain}, where data is easy to generate (such as in simulation), and a \textit{target domain}, where data is scarce or costly to collect (such as in the real world). The goal is to help models trained in the source domain perform well in the target domain despite this gap~\cite{zhao2020sim}. This is often achieved by aligning the feature spaces of the two domains using methods such as adversarial training, discrepancy minimization, or auxiliary reconstruction tasks~\cite{zhao2020sim}.
    
    For example, \citet{bousmalis2017unsupervised} propose an unsupervised domain adaptation method that transforms source-domain images to visually resemble target-domain images, while preserving task-relevant content. Their approach uses a GAN-based architecture to learn this pixel-level transformation without requiring paired examples across domains. \citet{jeong2020self} perform self-supervised domain adaptation by using the temporal structure in these data sequences (how observations change over time) to fine-tune the model’s visual perception module for real-world tasks.\index{domain adaptation|)}
    
    \item[\textbf{Meta learning}]\index{meta learning|(}
    Meta learning, i.e., ``learning to learn,'' aims to improve a model’s ability to quickly adapt to new tasks by leveraging experience gained across a distribution of related tasks~\cite{thrun1998learning}. Instead of training from scratch for each new task, a meta-learned model internalizes adaptation strategies, enabling efficient learning with minimal additional data or computation~\cite{zhao2020sim}.

    For example, in supervised settings, model-agnostic meta-learning enables image classifiers to quickly adapt to new classes with just a few labeled examples~\cite{finn2017model}. In meta reinforcement learning (MetaRL)~\cite{wang2016learning}, memory-based architectures (e.g., LSTM policies) capture temporal and task-specific patterns, enabling agents to quickly adapt to new real-world manipulation tasks after only a few trials by leveraging meta-learned experience from prior simulated tasks~\cite{pitkevich2024survey}.\index{meta learning|)}

    \item[\textbf{Robust RL}]\index{reinforcement learning!robust RL|(}
    Robust RL is a paradigm to explicitly take into account input disturbance as well as modeling errors~\cite{morimoto2005robust}. The objective is to learn policies that remain effective under adverse or unseen conditions, often modeled as worst-case scenarios~\cite{mankowitzrobust}. Techniques such as noise injection and adversarial training are commonly used to improve policy robustness~\cite{pitkevich2024survey}.

    For example, \citet{pinto2017robust} propose Robust Adversarial Reinforcement Learning (RARL), where a protagonist agent learns to perform tasks while withstanding disturbances from a learned adversary. RARL is trained in simulation, and improves robustness to model uncertainties and enables better generalization to unseen real-world variations like mass or friction changes.\index{reinforcement learning!robust RL|}

    \item[\textbf{Imitation learning}]
    Imitation learning aims to extract knowledge from human demonstrations or artificially created expert agents to replicate their behavior~\cite{zheng2022imitation}. It includes methods such as behavior cloning, where agents learn observation-to-action mappings directly from demonstrations, and inverse reinforcement learning, which infers a reward function that explains the expert’s behavior~\cite{zhao2020sim}. This approach offers stable rewards and enhances sim-to-real transfer by using human expertise to guide the learning process~\cite{pitkevich2024survey}.
    
    For example, \citet{wong2022error} use imitation learning to train visuo-motor policies for mobile manipulation tasks based on expert demonstrations in a simulated kitchen. By integrating an error detection module that prevents execution in unfamiliar states, it enhances policy robustness and reduces the risk of unsafe behavior when deployed in the real world.
\end{description}

\subsubsection{Use cases of sim-to-real in different domains}

\begin{description}
    \item[\textbf{Robotics}] The key problems are multitasking, precision, and safety. Robotics requires adaptable platforms capable of performing diverse tasks~\cite{fang2018multi}, and accuracy is crucial when performing fine manipulation or precision-based tasks~\cite{akinola2020learning}. For example, \citet{fang2018multi} address the domain shift problem, where models trained in simulation fail to generalize to real-world conditions due to differences in physics and perception. They develop a multi-task domain adaptation framework that uses domain-adversarial loss to transfer grasping capabilities from simulation to real robots, ensuring both multi-task capability and accuracy.
    
    Furthermore, robotics operating in physical environments face significant risks where operational failures can result in safety hazards~\cite{abbas2024safety}.
    
    For example, \citet{abbas2024safety} propose a framework that integrates traditional deep reinforcement learning with ISO 10218 and IEC 61508 functional safety standards. They develop and validate the framework in simulation, and transfer it to a real robotic cell via the sim-to-real approach, ensuring safer and more reliable robotic system deployment.
    
    These factors motivate researchers to develop methods that enable robust sensor integration and adaptive control techniques, bridging the gap between simulation and real-world performance~\cite{da2025survey}.

    \item[\textbf{Transportation}] Sim-to-real transfer occurs across diverse transportation scenarios, including multi-agent coordination, signal control, and lane keeping problems.
    
    Transportation systems need policies capable of managing dynamic environments with multiple interacting agents~\cite{yao2025comal}. For example, \citet{li2024s2r} address multi-agent coordination challenges in autonomous driving by developing methods to transfer collaborative policies from simulation to real-world multi-robot testbeds, demonstrating significant reduction in the sim-to-real gap through domain randomization. They also address deployment gaps such as localization errors from GPS inaccuracies and communication latency.
    
    Furthermore, traffic signal control systems require effective policy transfer to handle complex traffic patterns. \citet{da2023sim2real} tackle this challenge by employing grounded action transformation to bridge the domain gap between simulated and real traffic scenarios, enabling effective real-world actions.
    
    Additionally, lane-keeping systems require reliable visual perception performance to ensure safe navigation. For example, \citet{pahk2023effects} train dual contrastive learning networks to convert simulated visual inputs into realistic representations, mitigating the visual discrepancy between simulated and real-world driving conditions.
    
    \item[\textbf{Other domains}] In a broader context, the challenges extend to managing complex system dynamics, handling resource management, and continuously updating recommendation policies.
    
    In the building energy domain, systems require accurate modeling of complex dynamics. For example, \citet{fang2023transferability} develop a sim-to-real transfer learning framework that uses simulation datasets to enhance building energy prediction performance, systematically investigating the effects of building types, climate zones, and data volume on prediction accuracy across different transfer learning models.
    
    In edge-cloud computing, platforms face resource management challenges due to inaccuracy in the emulation of real computational infrastructure caused by abstractions in simulators. For example, \citet{tuli2022simtune} develop SimTune, a framework that uses low-fidelity surrogate models to update high-fidelity simulator parameters, improving simulation accuracy and enabling generalization to unknown edge-cloud configurations.
    
    In recommender systems, platforms must continuously adapt to evolving user preferences through effective policy transfer~\cite{chen2021survey}. For example, \citet{chen2023sim2rec} present Simulation-to-Recommendation (Sim2Rec), which uses multiple simulators to generate diverse user behavior patterns and trains a context-aware policy that can transfer to real-world environments by recognizing various user patterns and making optimal decisions based on inferred environment parameters.
\end{description}

\subsection{Additional challenges}
Apart from the sim-to-real transfer, a few additional challenges pose barriers to successful simulation of AI traning data. In the following, we review of few of these.

\subsubsection{Validation of simulated data}

\index{validation|(}
In practice, there is no standardized benchmark for assessing whether synthetic data is representative or useful~\cite{rujas2025synthetic}. As a result, validity assessments often highly rely on domain-specific criteria, which limits the comparability and generalizability of results. For example, image synthesis field uses the Inception Score~\cite{salimans2016improved}, while healthcare field relies on process-based clinical metrics~\cite{chen2019validity}.

Moreover, even commonly used evaluation techniques, such as comparing descriptive or summary statistics, can be misleading. Generated samples may exhibit similar descriptive statistics to real data while having fundamentally different data point distributions, or conversely, machine learning models trained on synthetic versus real data may demonstrate similar performance but have significant differences in underlying statistical or distributional characteristics~\cite{figueira2022survey}. As \citet{lautrup2024systematic} note, ``summary statistics can sometimes show a good result for the wrong reasons."\index{validation|)}

\subsubsection{Extra-functional concerns}
Beyond fidelity challenges, simulation-based synthetic data raises several extra-functional concerns, including safety, reliability, and security.

Safety is emphasized in reinforcement learning scenarios. For example, \citet{zhang2024automated} highlight safe simulation environments to ensure DRL agents can be trained without introducing risks during policy exploration. \citet{tubeuf2023increasing} note the need to integrate additional constraints to bound the agent’s action space for safety concerns.
Reliability is a particular concern in multi-access edge computing (also known as mobile edge computing)~\cite{dong2019deep} due to its ultra-low latency guarantees.
To ensure data security, e.g., by mitigating data leakage and tampering risks, \citet{liu2022digital} combine blockchains with data consistency authentication methods for data traceability and integrity.

\subsubsection{Privacy preservation}
In privacy-sensitive domains, such as healthcare and finance, where access to real data is often legally or ethically restricted, synthetic data offers a promising alternative~\cite{nikolenko2021synthetic}. However, even synthetic datasets may leak sensitive information about real individuals if not properly designed and evaluated. \citet{jordon2022synthetic} emphasize that privacy is inherently statistical, referring to how much information synthetic data reveals about real samples. Even theoretical guarantees, such as differential privacy may fail in practice if implementations are sloppy.
In addition, the trade-off between fidelity and privacy is a challenge. \citet{galloni2020novel} argue that any synthetic data generation approach should ensure a measurable level of privacy while maintaining utility and statistical similarity to real data. \citet{adams2025fidelity} show that differential privacy-enforced models significantly disrupted correlation structures, reducing data fidelity.

\section{Digital Twins for AI Training}\label{sec:dt4ai}

\index{digital twin|(}
Digital twins (DTs)~\cite{rasheed2020digital} are a promising technology to support AI training. This is due to the close, bi-directional coupling of DTs with their physical counterparts, and due to high fidelity of simulators found in DTs.
Here, we discuss how DTs can aid AI training.

\subsection{Digital twins}

Digital twins are high-fidelity, real-time virtual replicas of physical assets, referred to as the physical twin~\cite{kritzinger2018digital} or actual twin~\cite{tao2019digital}. 
Digital twins reflect the prevalent state of the physical twin and provide cost-effective and safe alternatives for interacting with it.
Collections of digital and physical twins are referred to as digital twin systems.

A unique characteristic of the digital twin system is the strong bi-directional coupling between the digital and physical counterparts. 
The digital twin controls the physical twin through computational reflection---the behavior exhibited by a reflective system based on a causal connection to the reflected system~\cite{maes1987concepts}. Such a causal connection is often established by the digital twin continuously processing sensor data from the physical twin. 
This bi-directional coupling enables the digital twin to reason about current and future states of the physical twin and control it.

At the core of a digital twin, a model of the physical system allows to simulate its future states and control the physical system accordingly. The model is typically maintained by regular updates by data advanced sensor technology, and constitutes the foundation of the services provided by the digital twin.
Such services are often enabled by simulators~\cite{boschert2016digital}.
Simulators are programs that encode the probabilistic mechanism that represents the real phenomenon and enact this probabilistic mechanism over a sufficiently long period of time to produce traces that describe the system~\cite{zeigler2018theory}.

To illustrate the above definition, here are some examples of systems that rely on digital twins.
In smart building systems, the building is the physical twin and is monitored and controlled through a digital twin to achieve optimal temperature~\cite{elmokhtari2022development}.
In robotic assembly systems for smart manufacturing, the manufacturing system acts as the physical twin and is observed and controlled through a cloud-based digital twin~\cite{touhid2023building}. 
In smart grids, digital twins monitor and optimize the power generator network, and conduct preventive maintenance actions to increase the time between failure~\cite{liu2026introduction}.
In automotive systems, digital twins help with executing resource-intensive computations on behalf of the vehicle~\cite{ramdhan2025engineering}.

Conversely, here are some examples of systems that are \textit{not} digital twins.
Closed-loop control is not a digital twin. In some research communities, closed-loop control is considered a form of a digital twin due to the presence of a control element. However, a controller does not facilitate the computational reflection. Models@run.time are not digital twins either. While Models@run.time~\cite{blair2009models} support computational reflection of physical systems, they are not capable of controlling physical systems.

Digital twins open new frontiers in AI training, as explained in the next section.\index{digital twin|)}

\subsection{The Case for Digital Twin-Enabled AI simulation}

The case for using digital twins for AI simulation boils down to two key arguments.
First, the emergence of digital twins has elevated the quality, fidelity, and performance of simulators, as simulators are first-class components of digital twins that enable sophisticated services such as real-time adaptation~\cite{tomin2020development}, predictive analytics~\cite{paredis2024coock}, and process control of complex cyber-biophysical systems~\cite{david2023digital}. These advanced capabilities require high-fidelity simulators that align well with AI training.
Second, through the close coupling between digital and physical components, digital twins enable purposeful experimentation with physical systems. When simulators lack knowledge for specific AI agent queries or contain outdated information, digital twins can collect samples from real physical settings to update simulation models, supporting more targeted and automated experimentation~\cite{liu2025ai}.

When digital twins are used for AI training, they support two types of activities: virtual training environments and data generation or labeling.

\begin{description}
    \item[\textbf{Virtual training environments}] The digital twin provides an interactive virtual environment in which AI agents can safely train through frequent interaction with the simulated environment. For example, \citet{shen2022deep} propose a digital twin-based deep reinforcement learning framework for unmanned aerial vehicles (UAVs), keeping the simulation model within the digital twin up-to-date. \citet{verner2018robot} use a digital twin to allow robots to safely explore and learn motion responses to unseen situations in a simulated environment.

    \item[\textbf{Data generation or labeling}] The digital twin generates labeled or controllable synthetic data to support offline AI training. For example, \citet{alexopoulos2020digital} develop a digital twin for manufacturing that can generate virtually created and labeled datasets to train AI agents. \citet{pun2023neural} present LightSim, which constructs lighting-aware digital twins to generate diverse, photorealistic driving videos under different lighting conditions. 
\end{description}

\subsection{The DT4AI framework}

\begin{figure}[h!]
    \centering
    \includegraphics[width=0.6\linewidth]{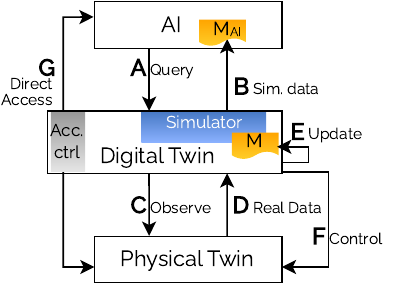}
    \caption{The DT4AI framework}
    \label{fig:framework}
\end{figure}

\index{digital twin|(}
To enable using digital twins for AI simulation, the DT4AI framework~\cite{liu2025ai}, shown in \figref{fig:framework}, situates digital twin components within the typical workflow of AI training data generation.
The framework consists of three main components. (i) \textit{AI}: the artificial intelligence agent under training. (ii) \textit{Digital Twin}: the high-fidelity virtual replica of the physical twin. (iii) \textit{Physical Twin}: the actual physical system.

\phantom{}

\textbf{AI training} involves interactions between the \textit{AI} agent and the \textit{Digital Twin}.

\begin{description}

    \item[\textbf{A: \textit{Query}}] Represents the request for data issued by the \textit{AI} component to the \textit{Digital Twin}. As shown in \tabref{tab:framework}, the \textit{Query} request can be either explicit (the \textit{AI} agent actively pulling data) or implicit (the \textit{Digital Twin} pushing data).
    
    \item[\textbf{B: \textit{Simulated data}}] The result of a simulation is a simulation trace, i.e., the data the \textit{AI} component receives in response to the \textit{Query}. The \textit{Digital Twin} is equipped with a (set of) model(s) \textit{M}, which serves the input to the \textit{Simulator}.
    
    The \textbf{A-B} training cycle can take either a batch or live format. In the former, the \textit{Trace volume} is big data; in the latter, the trace consists of small pieces of data.

\end{description}

\textbf{Data collection} involves interactions between the \textit{Digital} and the \textit{Physical Twin}.

\begin{description}
    \item[\textbf{C: \textit{Observe}}] The \textit{Digital Twin} is connected to the \textit{Physical Twin} through a data link and is able to passively observe or actively interrogate the \textit{Physical Twin}.

    \item[\textbf{D: \textit{Real data}}] Represents the data collected from the \textit{Physical Twin}. Depending on the type of the \textit{Observation}, \textit{Data} might be of low context, i.e., large volumes with low information value~\cite{david2024infonomics} (in case of passive observation); or of high context, i.e., smaller volumes of data in response to active experimentation. In situations when the \textit{Digital Twin} gets detached from the \textit{Physical Twin}, e.g., due to the retirement of the latter, data can be historical as well.
    
    As shown in \tabref{tab:framework}, the \textbf{C-D} Observe/Data cycle can be automated (scheduled by the \textit{Digital Twin}) or on-demand (based on the requests of the \textit{AI} or human operators).
    
    \item[\textbf{E: \textit{Update}}] After collecting data from the \textit{Physical Twin}, the model (\textit{M}) of the \textit{Digital Twin} needs to be updated in order to reflect the new data in simulations and transitively. This \textit{Update} can be achieved in a synchronous (blocking behavior but easier implementation) or asynchronous fashion (non-blocking behavior but more complex implementation, e.g., timeout and request obsolescence management). This step is known from works on digital twin evolution~\cite{david2023towards,michael2024smart} but it is seldom contextualized within purposeful experimentation.
\end{description}

\textbf{Control and access control} involve interactions between the \textit{Digital Twin} and the \textit{Physical Twin}.

\begin{description}
    \item[\textbf{F: \textit{Control}}] As customary, the \textit{Digital Twin} can control the \textit{Physical Twin} through the usual control links. As listed in \tabref{tab:framework}, control can be achieved \textit{in-place}, e.g., a learned policy on the digital side can govern the behavior of the physical system; or (parts of) the control logic can be \textit{deployed} onto the \textit{Physical Twin} for local control.

    \item[\textbf{G: \textit{Access control}}] The \textit{AI} component might interact with the \textit{Physical Twin} without the participation of the simulation facilities of the \textit{Digital Twin}. In these situations, the \textit{Digital Twin} provides \textit{Access control} to the \textit{Physical Twin}. 

\end{description}

The interactions (labeled with A--G) imply increasing complexity of AI training workflows.
For example, a basic interaction (A--B training cycle) describes a simple round-trip within the digital environment, such as training an AI agent using a sufficiently detailed and valid simulation model. However, interaction with the physical realm becomes necessary in some cases, adding complexity to the system (C--D observation cycle and E). This interaction is often limited to exploring desired states of the physical system, which require appropriate control (F). Finally, the AI agent may directly access the Physical Twin (G), but this is considered a rare and special scenario, thus appears last in the framework.

\begin{table*}
\centering
\caption{Variation points in the DT4AI framework}
\label{tab:framework}
\begin{tabular}{@{}ll@{}}
\toprule

\multicolumn{2}{c}{\textbf{AI training}} \\ 
\textbf{A} Query & \{Implicit, Explicit\} \\
\textbf{B} Sim. data volume & \{Big data , Small data\} \\
\textbf{A-B} Training fashion & \{Batch, Live\}\\

\midrule \multicolumn{2}{c}{\textbf{Data collection}} \\ 
\textbf{C} Observe & \{Passive observation, Active experimentation\}\\
\textbf{D} Data & \{Stationary historical data, Low-context data update, High-context data update\}\\
\textbf{C-D} Observe/Data trigger & \{Automated, On-demand\}\\
\textbf{E} Update synchronicity & \{Synchronous, Asynchronous\}\\

\midrule \multicolumn{2}{c}{\textbf{Control}} \\ 
\textbf{F} Control & \{In-place control, Deploy-and-Control\} \\

\bottomrule
\end{tabular}
\end{table*}

\index{digital twin|)}

\subsection{Typical instantiations} \label{sec:dt4ai-instantiations}

\begin{figure*}
    \centering    
    \begin{subfigure}{0.32\textwidth}
        \includegraphics[height=3.75cm]{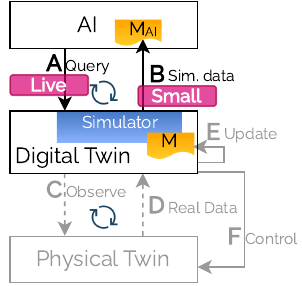}
        \caption{Reinforcement Learning}
        \label{fig:patterns-ai-rl}
    \end{subfigure}
    \hfill
    \begin{subfigure}{0.32\textwidth}
        \includegraphics[height=3.75cm]{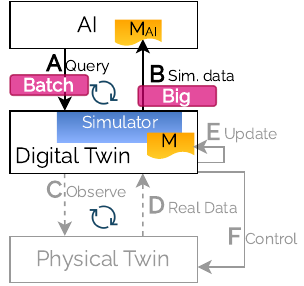}
        \caption{Deep Learning}
        \label{fig:patterns-ai-dl}
    \end{subfigure}
    \hfill
    \begin{subfigure}{0.32\textwidth}
        \includegraphics[height=3.75cm]{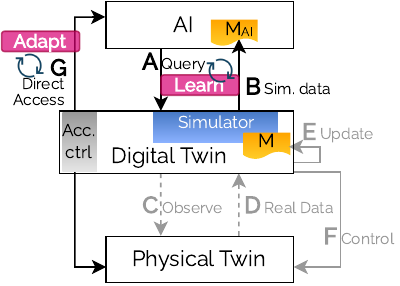}
        \caption{Transfer Learning}
        \label{fig:patterns-ai-tl}
    \end{subfigure}
    \caption{AI patterns (relevant components highlighted)}
    \label{fig:patterns-ai}
\end{figure*}

Instantiations of the DT4AI framework allow for different AI training workflows, as shown in \figref{fig:patterns-ai}.

\index{reinforcement learning (RL)|(}
Structurally, \textit{Reinforcement learning} (\figref{fig:patterns-ai-rl}) and \textit{Deep learning} (\figref{fig:patterns-ai-dl}) are identical. However, there are important differences in the interactions within the \textbf{A-B} learning cycle. 
\textit{Reinforcement learning} establishes a \textit{live} interaction, where the \textit{AI} issues frequent, short queries for \textit{small} amounts of simulated data. 
For example, \citet{cui2023digital} demonstrate how the DT continuously provides channel estimates and emulated rewards through live interactions to train and maintain a deep reinforcement learning (DRL) model in the virtual environment, where the DRL model can be continuously updated over time based on these frequent feedback exchanges.\index{reinforcement learning (RL)|)}

\index{deep learning|(}
In contrast, \textit{Deep learning} uses infrequent queries (often a singular one) to which the \textit{Digital twin} responds with \textit{big} data. 
For example, \citet{dong2019deep} illustrate how the DT generates various labeled training samples by computing energy consumption, delay, and packet loss probability across various network scenarios, enabling the deep neural network to be trained offline with extensive datasets in batch mode.\index{deep learning|)}

\index{transfer learning|(}
\textit{Transfer learning} (\figref{fig:patterns-ai-tl}) makes use of the \textit{Physical twin}, in which a DT is used as a proxy to the physical system for the AI agent to interact with~\cite{tubeuf2023increasing}. Here, the DT acts as a policy enforcer. After the \textit{learning} phase, the \textit{AI} interacts with the \textit{Physical twin} to \textit{adapt} the previously learned knowledge---either to adopt the knowledge to a changing environment or to mitigate sim-to-real threats~\cite{zhao2020sim}. 
In support of this process, the \textit{Digital Twin} ensures the necessary reliability, safety, and security measures~\cite{tubeuf2023increasing}.\index{transfer learning|)}

\subsection{How to use the framework}

The DT4AI framework is a conceptual framework, i.e., its purpose is to identify key concepts, components, and concerns in DT-enabled AI simulation, and map their relationships. Such high-level frameworks are of high utility during the conceptual design phase of complex systems, such as DTs for AI simulation. However, for technical design, stronger architectural foundations are required.

\index{digital twin!ISO 23247|(}
\citet{liu2025ai} provide a mapping on the ISO 23247 standard (``Digital twin framework for manufacturing'')~\cite{shao2021use}. This mapping onto an ISO-standard reference architecture provides researchers and practitioners with clear directives in their digital twin software development and standardization endeavors. The reader is referred to \citet[Section 7]{liu2025ai} for more details.
Although geared towards manufacturing systems, the ISO 23247 reference architecture has drawn increasing interest, e.g., in edge computing~\cite{kang2025edge} and its adaptation for new domains has already begun, e.g., in automotive systems~\cite{ramdhan2025engineering}, space debris detection~\cite{shtofenmakher2024adaptation}, and battery systems~\cite{cederbladh2024towards}.
These adaptations underline the utility of standardized architectures in DT engineering.

While the adoption of ISO 23247 is increasing, reference \textit{implementations} are still missing. It is, therefore, the adopters' responsibility to implement the standard properly.\index{digital twin!ISO 23247|)}

\section{Conclusion and Future Directions}\label{sec:conclusion}

As AI is becoming an increasingly larger part of modern socio-technical systems, efficient training and the related data concerns are of key importance. This chapter provided an introduction into how simulation-based synthetic data generation offers a compelling solution to the limitations of real-world data collection. We discussed various simulation techniques and workflows to support data generation, as well as key challenges, with a particular attention to the sim-to-real gap. Finally, we made a case for digital twins as the technological enablers of simulation-based AI training data generation, or, AI simulation.

Simulation accelerates the development of AI systems that would otherwise be hindered by issues of data scarcity, privacy, and safety. The alignment of simulation and AI technologies is getting recognized by companies and consulting firms, suggesting an increasing demand of state-of-the-art methods and technological solutions down the road.
At the same time, questions of fidelity, validity, and generalizability remain challenges to be addressed by prospective researchers and adopters.

Looking ahead, some key directions are the following. The maturation of digital twin technology positions it as the prime technological pathway to developing the next generation of adaptive and highly realistic AI simulation environments.
Advances in generative AI and foundation models figure to be valuable complements to traditional simulation by providing novel mechanisms to expand the diversity and representativeness of synthetic datasets.
Finally, interdisciplinary collaborations spanning computer science, AI, and various adopting domains will be critical in ensuring that simulation-based AI development is not only a technically apt solution, but also something that can be trusted in the development of AI systems.

In conclusion, simulation-based synthetic data generation represents both an apt solution to pressing data challenges and a paradigm shift in how AI systems are developed. Continued progress in AI simulation will open the door to systems that are smarter, safer, and better aligned with the needs of society.

\section*{Reflection and Exploration}

\begin{questype}{AI training}%
Think of an AI training task in a domain you are familiar with.
    \begin{enumerate}
        \item Explain which simulation methods in \secref{sec:sim4ai} would be suitable for your AI training purposes and why.
        \item Identify some of the challenges of AI simulation and their mitigation methods.
    \end{enumerate}
\end{questype}

\begin{questype}{Sim-to-real gap}%
Compare how the sim-to-real gap may manifest in different domains, e.g., robotics, transportation, and healthcare.
    \begin{enumerate}
        \item List one or two challenges that are unique to each domain.
        \item Identify the common challenges across all three domains.
    \end{enumerate}
\end{questype}

\begin{questype}{Domain randomization}%
Domain randomization aims to expose the model to enough simulated variability; however, ``over-randomization'' can degrade learning~\cite{li2024bridging}. When does domain randomization fail?
    \begin{enumerate}
        \item List two scenarios (domains or tasks) where domain randomization might harm performance, and explain why.
        \item Find a mitigation strategy for each scenario.
    \end{enumerate}
\end{questype}

\begin{questype}{AI simulation}%
List at least two advantages and two challenges of using digital twins for AI simulation.
\end{questype}

\begin{exploration}{Further exploration}%
Choose one peer-reviewed study that uses a digital twin to simulate data for AI training. For reference, you can follow ``AI Simulation by Digital Twins: Systematic Survey, Reference Framework, and Mapping to a Standardized Architecture''by \citet{liu2025ai} or another systematic review.
    \begin{enumerate}
        \item Identify how the study's case corresponds to the DT4AI framework.
        \item Identify which instantiation of the framework best describes the study.
        \item Briefly list a key benefit and a key limitation of using a digital twin in this specific case.
    \end{enumerate}
\end{exploration}

\bibliographystyle{plainnat}
\bibliography{references}
\end{document}